\documentclass[journal]{IEEEtran}
\usepackage{graphicx}
\usepackage{times}
\usepackage{epsfig}
\usepackage{graphicx}
\usepackage{amsmath}
\usepackage{amssymb}
\usepackage{bm}
\usepackage{amsfonts}
\usepackage{amssymb}
\usepackage{graphicx}
\usepackage{subfigure}
\usepackage{multirow}
\usepackage{booktabs}
\usepackage{diagbox}
\usepackage{makecell}
\usepackage{textcomp}

\usepackage[linesnumbered,ruled]{algorithm2e}
\usepackage[pagebackref=true,breaklinks=true,letterpaper=true,colorlinks,bookmarks=false]{hyperref}

\ifCLASSINFOpdf
\else
\fi
\begin{document}
%
\title{Unsupervised Person Re-identification: \\Clustering and Fine-tuning}
%
%
%

\author{Hehe Fan, \and  Liang Zheng and \and Yi Yang
\thanks{H. Fan, L. Zheng and Y. Yang are with the Center for Artificial Intelligence, University of Technology Sydney,  Australia. Email: hehe.fan@student.uts.edu.au, liang.zheng@uts.edu.au, yi.yang@uts.edu.au.}}

\maketitle

\begin{abstract}
The superiority of deeply learned pedestrian representations has been reported in very recent literature of person re-identification (re-ID). 
	In this paper, we consider the more pragmatic issue of learning a deep feature with no or only a few labels. We propose a progressive unsupervised learning (PUL) method to transfer pretrained deep representations to unseen domains. Our method is easy to implement and can be viewed as an effective baseline for unsupervised re-ID feature learning. Specifically, PUL iterates between 1) pedestrian clustering and 2) fine-tuning of the convolutional neural network (CNN) to improve the original model trained on the irrelevant labeled dataset. Since the clustering results can be very noisy, we add a selection operation between the clustering and fine-tuning. At the beginning when the model is weak, CNN is fine-tuned on a small amount of reliable examples which locate near to cluster centroids in the feature space.
	As the model becomes stronger in subsequent iterations, more images are being adaptively selected as CNN training samples. Progressively, pedestrian clustering and the CNN model are improved simultaneously until algorithm convergence. This process is naturally formulated as self-paced learning.
    We then point out promising directions that may lead to further improvement. Extensive experiments on three large-scale re-ID datasets demonstrate that PUL outputs discriminative features that improve the re-ID accuracy.
    Our code has been released at \url{https://github.com/hehefan/Unsupervised-Person-Re-identification-Clustering-and-Fine-tuning}.
\end{abstract}


%
\IEEEpeerreviewmaketitle

\section{Introduction}
\label{sec:intro}
%
%
%
%
\IEEEPARstart{T}{his} paper discusses person re-identification (re-ID), a task which provides critical insights whether a person re-appears in another camera after being spotted in one \cite{DBLP:journals/corr/ZhengYH16}. Recent works view it as a search problem aiming to retrieving the relevant images to the top ranks \cite{DBLP:conf/iccv/ZhengSTWWT15,DBLP:journals/corr/ZhengZY17}.

Large-scale learning methods based on the convolutional neural network (CNN) constitute the recent advances in this field. Broadly speaking, current deep learning methods can be categorized into two classes, \emph{i.e.,} similarity learning and representation learning. For the former, the training input can be image pairs \cite{varior2016gated}, triplets \cite{liu2016end}, or quadruplets \cite{chen2017beyond}. These methods directly output the similarity score of two input images by focusing on their distinct body patches, without an explicit feature extraction process. This line of methods are less efficient during testing. On the other hand, when learning feature representations, contrastive loss or softmax loss are most commonly employed. For these methods, there is an explicit feature extraction process for the query and gallery images, which is learned to discriminate among a pool of identities or among image pairs/triplets \cite{cheng2016person,DBLP:journals/corr/HermansBL17}. Then the re-ID procedure is rather like a retrieval problem under some commonly used similarity measurement such as Euclidean distance. Due to its efficiency in large-scale galleries and the potentials in further acceleration, the deep feature learning strategy has become more popular \cite{DBLP:journals/corr/ZhengZY17,DBLP:journals/corr/ZhengYH16,DBLP:journals/corr/HermansBL17,xiao2016learning}. Therefore, we adopt the classification model proposed in \cite{DBLP:journals/corr/ZhengYH16,xiao2016learning} for feature learning. 

\begin{figure*}
\centering
\includegraphics[width=0.92\textwidth]{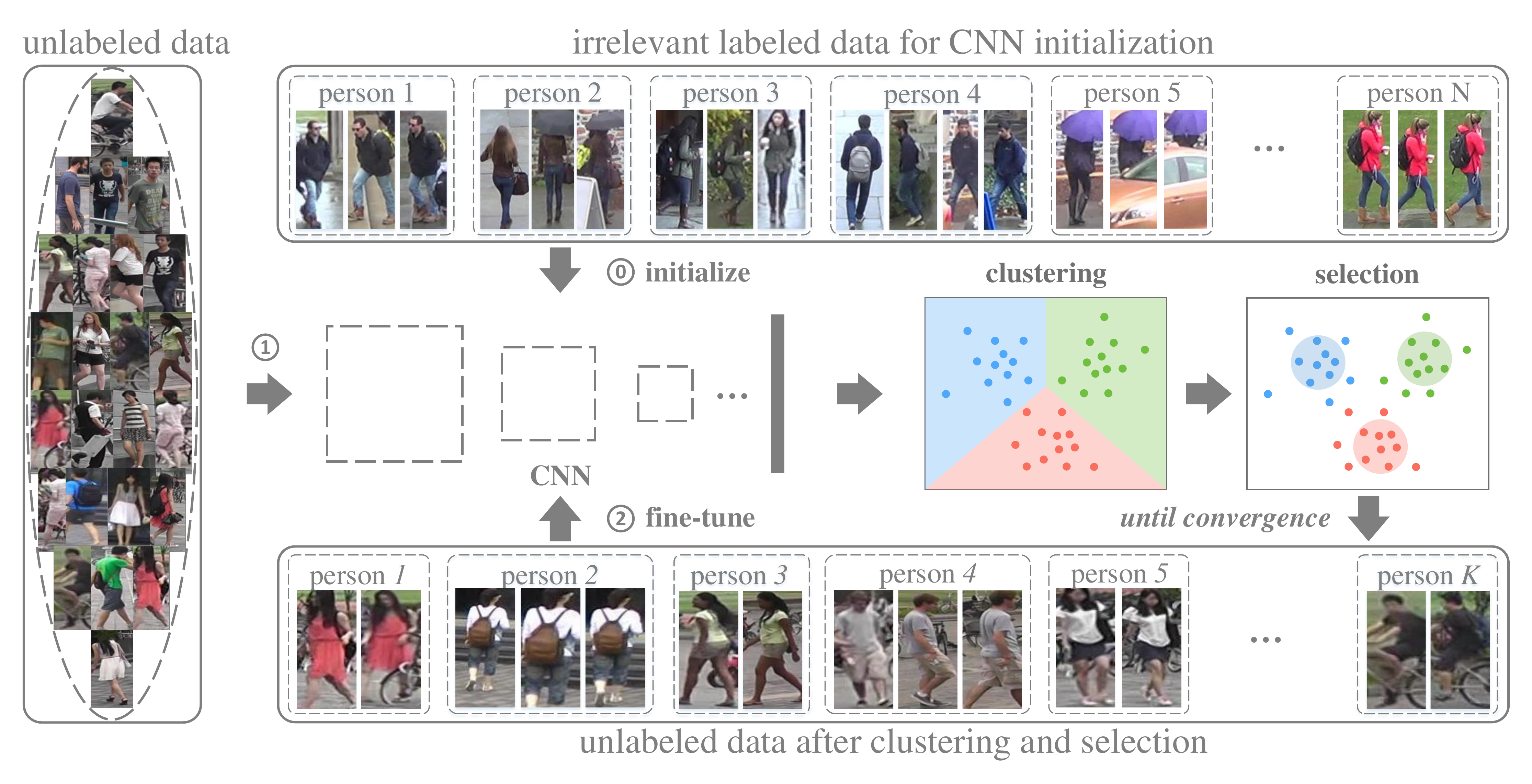}
\caption{Illustration of the PUL framework. In step 0, we initialize CNN on an irrelevant labeled dataset. Then we go into the iterations. During each iteration, we 1) extract CNN features for the unlabeled dataset, and perform clustering and sample selection, and 2) fine-tune the CNN model using the selected samples. Each cluster denotes a person.}
\label{Fig:Framework}
\end{figure*}

Despite the impressive progress achieved by the deep learning methods \cite{DBLP:journals/corr/ZhengZY17,DBLP:journals/corr/LinZZWY17,DBLP:journals/corr/HermansBL17}, most of these methods focus on re-ID with sufficient training data in the same environment. 
In fact, the lack of labeled training data under a new environment has been addressed by many previous works by resorting to unsupervised or semi-supervised learning \cite{DBLP:conf/eccv/KodirovXFG16,peng2016unsupervised,ma2017person,yang2017unsupervised,DBLP:journals/corr/YangWHTR17}. These works typically deal with the relatively small datasets \cite{gray2007evaluating,wang2014person} and to our knowledge, none of these works are integrated into the deep learning framework that has been proven as the state of the art under the large-scale datasets \cite{DBLP:conf/iccv/ZhengSTWWT15,DBLP:journals/corr/HermansBL17,sun2017svdnet,geng2016deep}. 
In fact, the problem of high annotation cost is more severe for deep learning based methods due to the intensive demand of data. So under this circumstances, we make initial attempts by developing a easy-to-implement method for unsupervised deep representation learning in the re-ID community. Our method can be easily extended to the semi-supervised setting by annotating a small amount of bounding boxes. We also notice some works on human-in-the-loop learning \cite{wang2016human,liu2013pop}, introducing labels by human feedback. We envision our work is complementary to this line of methods.  

In this paper, we aim at finding ``better'' feature representations without labeled training data to adapt to the unlabeled dataset. Named ``progressive unsupervised learning'' (PUL), our method is an iterative process and each iteration consists of two components (shown in Fig. \ref{Fig:Framework}). 1) We deploy the model pre-trained on some external labeled data to extract features of images from the unlabeled training set. Standard $k$-means clustering is performed on the CNN features, followed by a selection operation to choose reliable samples. 2) The original model is fine-tuned using the selected training samples. We then use this newborn model to extract features and begin another iteration of training. At the beginning, when the model is weak, the proposed approach learns from a small amount of reliable examples, which are near to cluster centroids in the feature space, to avoid getting stuck at a bad optimum or oscillating. As the model becomes stronger in subsequent iterations, more data instances are adaptively selected and exploited as training examples. By this progressive method, the model and clustering are trained and rectified alternately. This process can also be called self-paced learning \cite{DBLP:conf/nips/KumarPK10}.

The contributions of this paper are three-fold: 

1) Among the early efforts, we propose an easy-to-implement unsupervised deep learning framework for large-scale person re-ID. 

2) We are the first to discover the underlying relation between clustering and self-paced learning and integrate the self-paced learning paradigm into the unsupervised learning regime. This learning strategy facilitates PUL to extract faithful knowledge from highly unreliable clustering results in learning.  

3) Extensive experiments on three large-scale datasets indicate that our method noticeably improves the re-ID accuracy in the cross-dataset evaluation.

 

\section{Related Works}

\subsection{Deeply Learned Person Representations}
Deep learning models have been popular since Krizhevsky \textit{et al.} \cite{DBLP:conf/nips/KrizhevskySH12} won ILSVRC’12 by a large margin. Deeply learned person representations are producing state-of-the-art re-ID performance recently. The first two works in re-ID to use deep learning were \cite{DBLP:conf/icpr/YiLLL14} and \cite{DBLP:conf/cvpr/LiZXW14}. Generally speaking, from the perspective of model,  classification models as used in image classification \cite{DBLP:conf/nips/KrizhevskySH12} and siamese models that use image pairs \cite{DBLP:conf/eccv/RadenovicTC16} or triplets \cite{DBLP:conf/cvpr/SchroffKP15} are two types of CNN models that are commonly employed in re-ID. At the beginning when the training datasets are not big enough, such as VIPeR \cite{DBLP:conf/eccv/GrayT08} that provides only two images for each identity, the siamese model dominates re-ID community. As the scale of re-ID dataset becomes large, such as Market-1501 \cite{DBLP:conf/iccv/ZhengSTWWT15}, the classification model is widely employed. A survey of re-ID can be viewed in \cite{DBLP:journals/corr/ZhengYH16}.

On the other hand, comparing with deep similarity learning methods \cite{varior2016gated,liu2016end,chen2017beyond}, the representation learning methods are more extendable for large galleries. For example, Hermans \textit{et al.} \cite{DBLP:journals/corr/HermansBL17} used an efficient variant of the triplet loss based on the distance between samples. Another popular choice consists in training an identification network and extracting the intermediate output as a discriminative embedding \cite{DBLP:journals/corr/ZhengYH16,xiao2016learning,DBLP:conf/eccv/ZhengBSWSWT16,zheng2017pose}. For example, Xiao \emph{et al.} \cite{xiao2017joint} propose an online instance matching loss to address the problem of having only few training samples each class.  Lin \textit{et al.} \cite{DBLP:journals/corr/LinZZWY17} combine attribute classification losses and the re-ID loss for embedding learning. In order to exploit more training data, Zheng \textit{et al.} \cite{DBLP:journals/corr/ZhengZY17} employ the generative adversarial network to generate samples which are expected to generate uniform prediction probabilities in the softmax layer. In this paper, we adopt the baseline identification model, named ``ID-discriminative Embedding'' (IDE) in \cite{DBLP:conf/eccv/ZhengBSWSWT16}.

\subsection{Unsupervised Person Re-ID}
\label{sub:related-unsupervised}
Although deeply learned person representations have achieved significant progress by supervised learning methods, less attention is paid on unsupervised learning for re-ID. Kodirov \textit{et al.} \cite{DBLP:conf/eccv/KodirovXFG16} utilize a graph regularized dictionary learning to
capture discriminative cues for cross-view identity matching. Yang \textit{et al.} \cite{yang2017unsupervised} propose a weighted linear coding method to learn multi-level descriptors from raw pixel data in an unsupervised manner. Wang \textit{et al.} \cite{wang2016towards} use a kernel subspace learning model to learn cross-view identity-specific information from unlabeled data. Peng \textit{et al.} \cite{peng2016unsupervised} propose a multi-task dictionary learning model to transfer a view-invariant representation from a number of existing labeled source datasets to an unlabeled target dataset. Ma \textit{et al.} \cite{ma2017person} utilize image-sequence information to fuse different descriptions. These methods focus on small-scale datasets and do not adopt deep features.

Another choice in unsupervised re-ID consists in deploying hand-crafted features directly. Several effective descriptors have been developed in recent years. In the earlier works, Farenzena \cite{DBLP:conf/cvpr/FarenzenaBPMC10} \textit{et al.} use the weighted color histogram, the maximally stable color regions, and the recurrent high-structured patches to segment the pedestrian foreground from the background. Gray and Tao \cite{DBLP:conf/eccv/GrayT08} use 8 color channels and 21 texture filters on the luminance channel, and the pedestrian is partitioned into horizontal stripes. Recently, Zhao \textit{et al.} \cite{DBLP:conf/cvpr/ZhaoOW13,DBLP:conf/iccv/ZhaoOW13,DBLP:conf/cvpr/ZhaoOW14} use the 32-dim LAB color histogram and the 128-dim SIFT descriptor are extracted from each 10 $\times$ 10 patch densely sampled with a step size of 5 pixels. Liao \textit{et al.} \cite{DBLP:conf/cvpr/LiaoHZL15} propose the local maximal occurrence (LOMO) descriptor, which includes the color and SILTP histograms. LOMO is later employed by \cite{DBLP:conf/cvpr/ZhangXG16}, \cite{DBLP:conf/cvpr/ZhangLLIR16} and a similar set of features is used by Chen \textit{et al.} \cite{DBLP:conf/cvpr/ChenYCZ16}. Zheng \textit{et al.} \cite{DBLP:conf/iccv/ZhengSTWWT15} propose extracting the 11-dim color names descriptor for each local patch, and aggregating them into a global vector through a Bag-of-Words  model.

\subsection{Self-paced Learning}
In the human learning process, knowledge is imparted in the form of curriculum and organized from easy to difficult. Inspired by this process, the theory of Curriculum Learning (CL) \cite{DBLP:conf/icml/BengioLCW09} is proposed to accelerate the speed of convergence of the training process to a minimum. The main challenge of CL is that it requires a known separation to indicate whether a sample in a given training dataset is easy or hard. To alleviate this deficiency, self-paced learning \cite{DBLP:conf/nips/KumarPK10,DBLP:conf/nips/JiangMYLSH14,ma2017spacotrain} embeds easiness identification into model learning. The learning principle behind SPL is to prevent latent variable models from getting stuck in a bad local optimum or oscillating. Then SPL is widely used in weakly-supervised and semi-supervised learning. For example, Zhang \textit{et al.} \cite{DBLP:journals/pami/ZhangMH17} integrate self-paced learning into multi-instance learning for co-Saliency detection.

In this paper, we utilize SPL to select reliable samples to fine tune the original model. If we consider the labels of the samples from the unlabeled dataset as latent variables, our method also belongs to latent variable model. Therefore, it is necessary to avoid getting stuck in bad optimums or oscillating using SPL.

\section{Progressive Unsupervised Learning}
\label{progressive}

In this section, we present our progressive unsupervised learning framework for re-ID in detail. The proposed approach begins with a basic convolutional neural network, such as the ResNet-50 \cite{DBLP:conf/cvpr/HeZRS16} network trained on the \textit{ImageNet}. Firstly, we fine tune the basic model by an irrelevant labeled dataset, which is stored as the original model. This step is also called original model initialization. Secondly, the original model is used to extract feature for the samples from the unlabeled dataset. The data instances are clustered and selected to generate a reliable training set. Thirdly, the original model is fine tuned by this reliable training set. Lastly, we use the new model to extract feature, and the second and third steps are repeated until the scale of the reliable training set becomes stable. The progressive unsupervised learning framework is demonstrated in Fig \ref{Fig:Framework}.

\subsection{Formulation}
\label{subsec:formulation}
Suppose the unlabeled dataset contains $N$ cropped person images $\{x_i\}_{i=1}^N$ with $K$ identities. Since the data is not labeled, we consider the identities $\{y_i\}_{i=1}^N$ of $\{x_i\}_{i=1}^N$ as latent variables. Therefore, our PIL belongs to latent variable model.

Let $v_i$ be the selection indicator of sample $x_i$. If $v_i$ equals to 1, $x_i$ is selected as a reliable sample; otherwise, $x_i$ is discarded when fine tuning the original model. We further denote $\mathbf{v} = [v_1, \ldots, v_N]$ as the selection indication vector and $\mathbf{y} = [y_1, \ldots, y_N] \in \{1, \ldots, K\}^N$ as the label vector for $\{x_i\}_{i=1}^N$. The CNN model is as denoted $\phi(\cdot;\boldsymbol{\theta})$, which is parameterized by $\boldsymbol{\theta}$. After $\phi(\cdot;\boldsymbol{\theta})$, a one-dimensional feature vector is generated, which is fed into a classifier parameterized by $\mathbf{w}$. The $\boldsymbol{\theta}$ and $\mathbf{w}$ are optimized simultaneously under a classification model. 

We formulate our idea as optimizing the following three problems alternately:
\begin{equation}\label{eq1}
\underset{\mathbf{y}, \mathbf{c}_1, ..., \mathbf{c}_K}{\min}\sum_{k=1}^K\sum_{y_i=k} \Vert \phi(x_i;\boldsymbol{\theta}) - \mathbf{c}_k \Vert^2.  
\end{equation}
\begin{equation}\label{eq2}
\begin{split}
\underset{\mathbf{v}}{\min} \sum_{k=1}^K\sum_{y_i=k}   v_i & \Vert \phi(x_i;\boldsymbol{\theta}) - \mathbf{c}_k \Vert - \lambda \Vert \mathbf{v} \Vert_1, \\
s.t. \ v_i \in \{ & 0, 1\}; \sum_{y_i=k}v_i \geq 1, \forall k.
\end{split}
\end{equation}
\begin{equation}\label{eq3}
\underset{\boldsymbol{\theta},\mathbf{w}}{\min} \sum\limits_{i=1}^N v_i \mathcal{L}(y_i, \phi(x_i; \boldsymbol{\theta}); \mathbf{w}),
\end{equation}
where $\mathbf{c}_k$ is the mean of points $\{\phi(x_i;\boldsymbol{\theta})\}_{y_i=k}$ in the feature space, $\lambda$ is a positive parameter and $\mathcal{L}$ denotes a loss function for classification. 

The first step (Eq. \ref{eq1}) is to infer the images' labels $\mathbf{y}$ by the results of the standard $k$-means clustering. Since the noisy clustering results will make this latent variable model prone to getting stuck in a bad local optimum or oscillating, it is inappropriate to directly use them to fine-tune the CNN model. 

To alleviate the above problem, rather than considering all samples simultaneously, our PUL selects the training data in a meaningful order that facilitates learning. The order of the samples is determined by how reliable they are. The second step (Eq. \ref{eq2}) is to select reliable data instances that are close enough to the cluster centroids. We formulate this operation as self-paced learning \cite{DBLP:conf/nips/KumarPK10}. The ``easiness'' self-paced learning is equal to ``reliability'' in PUL, and in essence, it depends on the Euclidean distance in clustering. The sample closer to the centroid is considered as an easy and reliable training instance.  The constraint $\sum_{y_i=k}v_i \geq 1$ can guarantee that each cluster contains at least one reliable sample. 

The third step (Eq. \ref{eq3}) is to fine-tune the model by the clustering and selection results from Eq. \ref{eq1} and Eq. \ref{eq2}. Generally speaking, there are three types of loss functions for CNN models are commonly employed in the community. The first one is the classification method, in which a classifier is added at the end of model $\phi(x;\boldsymbol{\theta})$ as a fully-connected layer. When $v_i$ equals to 0, the loss will not be calculated, which means the corresponding sample will not be used to fine tune the original model. Eq. \ref{eq3} can be replaced with the contrastive
loss \cite{DBLP:conf/eccv/RadenovicTC16} or triplet loss \cite{DBLP:conf/cvpr/SchroffKP15}. In this paper, we mainly focus on the classification model.

\begin{figure}
\includegraphics[width=0.45\textwidth]{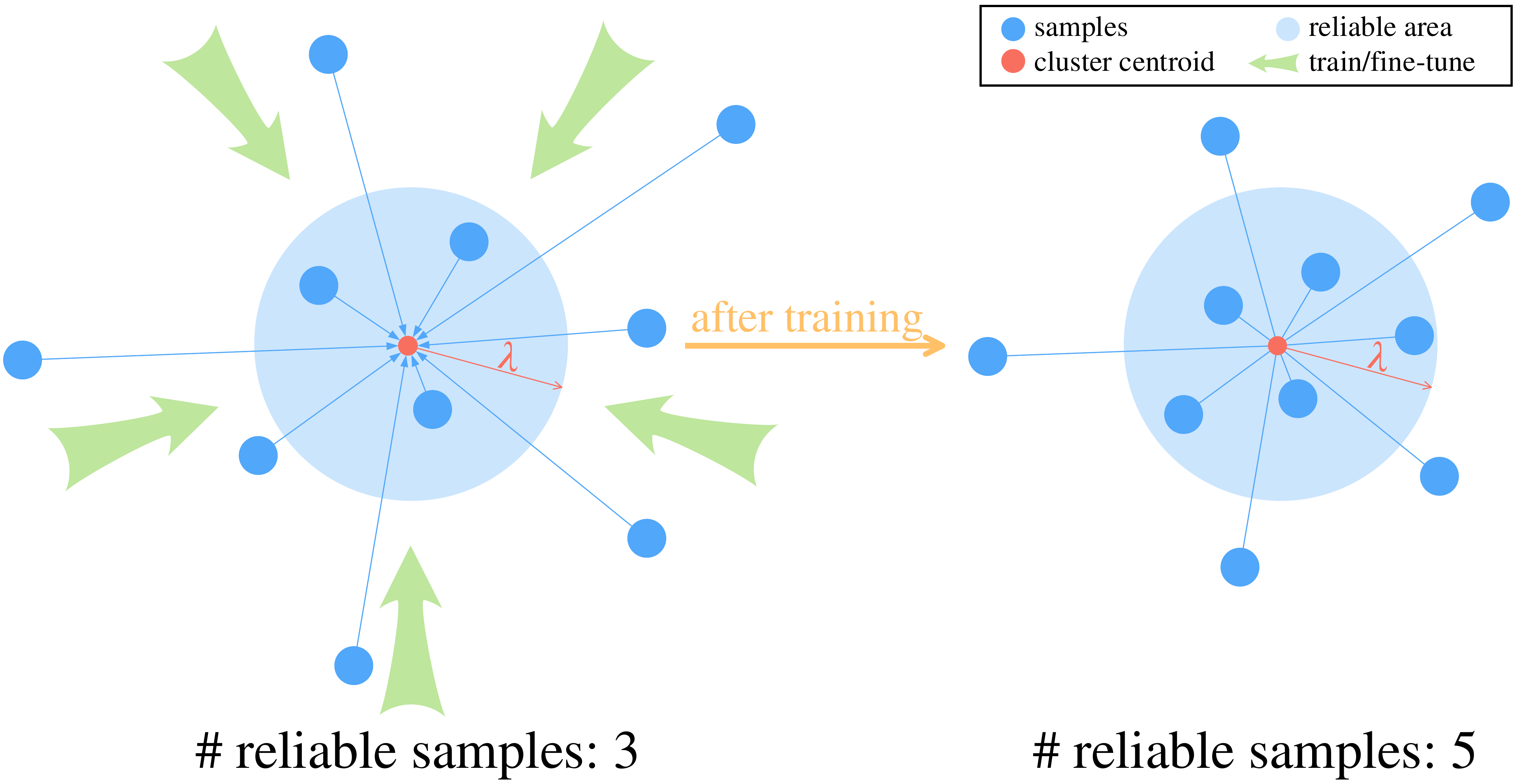}
\caption{Illustration of the learning process. Reliable samples covered in the large blue circle are used for CNN fine-tuning. As the CNN model gets fine-tuned, more challenging training data can be mined.} 
\label{Fig:Reliability}
\end{figure}

\begin{figure*}[t]
\centering
\includegraphics[width=1.0\textwidth]{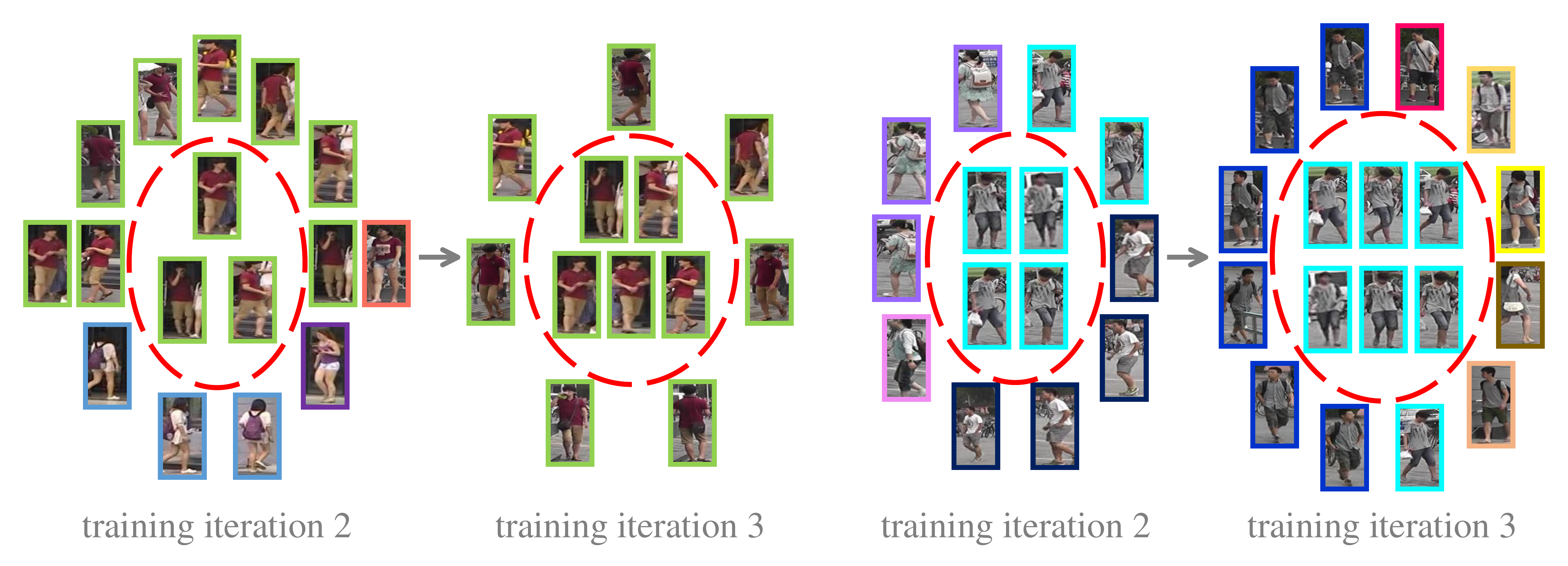}
\caption{Two visual examples of the working mechanism of progressive unsupervised learning (PUL). Each color denotes a distinct ID and the red circles represent reliable areas. From training iteration 2 to training iteration 3, more samples are selected reliable samples. In PUL, when the number of selected samples is saturated, the algorithm reaches convergence.}
\label{Fig:Example}
\end{figure*}

Eq. \ref{eq1}, Eq. \ref{eq2} and Eq. \ref{eq3} constitute a training iteration. As training goes on, we obtain more discriminative CNN models which facilitate the clustering process with smaller clustering errors. That is, with the optimization of Eq. \ref{eq1} and Eq. \ref{eq3}, the clustering error $\Vert \phi(x_i;\boldsymbol{\theta}) - \mathbf{c}_k \Vert^2$ will become smaller and smaller. From the perspective of feature representations, the optimization of Eq. \ref{eq3} will change the distribution of data instances in the feature space, which makes the images corresponding to the same person get closer to cluster centroid.
This leads to an ``self-paced'' scheme in which more selection indicators $v_i$ are expected to become 1 in order to minimize \ref{eq2}. Therefore, more and more examples are selected into the training set, until no more reliable data instances can be added. Fig  \ref{Fig:Reliability} illustrates how the training set changes during optimization. We also list two visual examples of this ``self-paced'' process from the following experiments in Fig \ref{Fig:Example}.

\subsection{Optimization Procedure}
\label{sub:optimization}
The optimization Eq. \ref{eq1}, Eq. \ref{eq2} and Eq. \ref{eq3} can be solved in an alternate manner. 

\textit{1) Optimize $\mathbf{y}$ and $\mathbf{c}_1, ..., \mathbf{c}_K$ when $\mathbf{\theta}$ and $\mathbf{v}$ are fixed.} In this step, the optimization problem reduces to the classic $k$-means clustering and can be easily solved. 

\textit{2) Optimize $\mathbf{v}$ when $\boldsymbol{\theta}$, $\mathbf{y}$ and $\mathbf{c}_1, ..., \mathbf{c}_K$ are fixed.} The goal of this step is to select reliable data instances to fine tune the CNN model with the clustering results $\mathbf{y}$ in Eq. \ref{eq1}. To minimize the first component of Eq. \ref{eq2}, all selection indicators $\{v_i\}$ can be trivially set to be 0, which means all shots are not selected. However, the second component $l_1$-norm $- \Vert \mathbf{v} \Vert_1$ monotonically increases as the number of selected shots decreases. So it encourages all selection indicators to be 1. By making a balance between these two components, the objective function aims to training the CNN model with reliable shots. A sample will be selected if the distance of its feature to its corresponding cluster centroid is smaller than $\lambda$. Here, $\lambda$ is a threshold to control the reliable sample selection. 

\textit{3) Optimize $\boldsymbol{\theta}$ and $\mathbf{w}$ when $\mathbf{v}$, $\mathbf{y}$ and $\mathbf{c}_1, ..., \mathbf{c}_K$ are fixed.} At the beginning of each iteration, the $\mathbf{w}$ is initialized randomly. 


In practice, we use cosine to measure the similarity between two feature vectors. The optimization algorithm of PUL is presented in Alg. \ref{alg:ProgressiveUnsupervisedLearning}. To guarantee each cluster contains at least one reliable sample, we choose the nearest feature to the corresponding centroid as the center of the cluster. In the algorithm, samples with $\mathbf{f}_i \cdot \mathbf{f}_k > \lambda$ will be selected for fine-tuning ($v_i=1$); otherwise,  sample will not be selected ($v_i=0$). When the number of selected reliable samples is saturated, the algorithm will converge.

\begin{algorithm}[t]
\SetAlgoLined
\SetKwInOut{Input}{Input}
\SetKwInOut{Output}{Output}
\SetKwInput{Initialization}{Initialization}
\caption{Progressive Unsupervised Learning}\label{alg:ProgressiveUnsupervisedLearning}
\Input{Unlabeled data $\{x_i\}_{i=1}^N$; \\
	   Reliability threshold $\lambda$; \\
       Number of clusters $K$; \\
       Original model $\phi(\cdot, \boldsymbol\theta_o)$.}
\Output{Model $\phi(\cdot, \boldsymbol\theta_t)$.}
\Initialization{$\boldsymbol\theta_o \rightarrow \boldsymbol\theta_t$.} 
\While{not convergence}{
	randomly initialize $\mathbf{w}$\;
	extract feature: $\mathbf{f}_i = \phi(x_i, \boldsymbol\theta_t)$ for all $i$\;
    $k$-means: update $\{y_i\}_{i=1}^N$ and $\{\mathbf{c}_k\}_{k=1}^K$\;
    select center features: $\{\mathbf{c}_k\}_{k=1}^K \rightarrow \{\mathbf{f}_k\}_{k=1}^K$\;
    $l_2$-normalize $\{\mathbf{f}_i\}_{i=1}^N$\;
    \For{$k = 1$ to $K$}{\label{alg:selection-1}
    	\For{$i = 1$ to $N$}{\label{alg:selection-3}
        	calculate inner product $\delta_i = \mathbf{f}_i \cdot \mathbf{f}_k$\;
            \eIf{$\delta_i>\lambda$}{
                $v_i \leftarrow 1$; // selected
            }{
                $v_i \leftarrow 0$; // removed
            }
        }\label{alg:selection-4}
    }\label{alg:selection-2}
    fine-tune $<\phi(\cdot, \boldsymbol\theta_o), \mathbf{w}>$ with selected samples $\rightarrow \boldsymbol\theta_t$\;
}
\end{algorithm}

\subsection{Semi-supervised PUL}
\label{sub:semi-supervised}

The PUL framework can be easily extended to semi-supervised learning. To integrate the labeled data into the training process, we just need to directly add the labeled data into the selected reliable training set in every iteration. 

The semi-supervised PUL is a general case of our model. When all training data is labeled, since there is no need to infer labels or select reliable samples, the semi-supervised PUL degenerates supervised learning; when no labeled data is available, the method has to rely on clustering and self-paced learning to obtain reliable training data.

\section{Experiments}
\subsection{Datasets and Settings}

We mainly evaluate the proposed method on DukeMTMC-reID \cite{DBLP:journals/corr/ZhengZY17} and Market-1501 \cite{DBLP:conf/iccv/ZhengSTWWT15}, because they are two relatively large datasets with multiple cameras. We also report results on CUHK03 \cite{DBLP:conf/cvpr/LiZXW14}, which is a relatively small dataset. We do not use MARS \cite{DBLP:conf/eccv/ZhengBSWSWT16}, another large-scale dataset, because it is an extension of Market-1501 and hence has a similar data distribution with Market-1501.

\textbf{DukeMTMC-reID} is a subset of the pedestrian tracking dataset DukeMTMC \cite{DBLP:conf/eccv/RistaniSZCT16}. This dataset contains 36,411 images with 1,812 identities captured from 8 different view points. Following Market-1501, the dataset is also split into three parts: 16,522 images with 702 identities for training, 17,661 images with 1,110 identities in gallery, and another 2,228 images with 702 identities in the gallery for query. Images in this dataset vary in size. For simplicity, we use ``Duke'' to represent this dataset.

\begin{table}[t]
\setlength{\tabcolsep}{3.15pt}
\caption{Statistics of Duke \cite{DBLP:journals/corr/ZhengZY17}, Market \cite{DBLP:conf/iccv/ZhengSTWWT15} and CUHK03 \cite{DBLP:conf/cvpr/LiZXW14} datasets. Note that we use the train/test protocol proposed in \cite{DBLP:journals/corr/ZhongZCL17} for CUHK03. For Market and CUHK03, the query and gallery share the same IDs. However, for Duke, except for the shared 702 IDs, the gallery contains another 408 IDs.}
\label{tabel:dataset}
\centering
\vspace{1.0 ex}
\begin{tabular}{l|c|cc|cc|cc}
\hline
\multirow{3}{*}{datasets} & \multirow{3}{*}{\# cams} & \multicolumn{2}{c|}{training}                          & \multicolumn{4}{c}{test}                                   \\ \cline{3-8}
						  &                          & \multirow{2}{*}{\# IDs}  & \multirow{2}{*}{\# images}  & \multicolumn{2}{c|}{query}  & \multicolumn{2}{c}{gallery}  \\ \cline{5-8}
                          &                          &                          &                             & \# IDs   & \# images       & \# IDs   & \# images          \\ \hline\hline
Duke                      & 8                        & 702                      & 16,522                      & 702      & 2,228           & 1,110    & 17,661             \\
Market                    & 6                        & 751                      & 12,936                      & 750      & 3,368           & 750      & 19,732             \\
CUHK03                    & 2                        & 767                      & 7,365                       & 700      & 1,400           & 700      & 5,332              \\ \hline
\end{tabular}
\end{table}

\begin{table*}[t]
\setlength{\tabcolsep}{4.0pt}
\caption{Person re-ID accuracy (rank-1,5,10,20 precision and mAP) of two methods: the baseline (fine-tuned ResNet-50 on one of datasets and tested on another one) and PUL. Note that PUL work on unsupervised settings which need the baseline model for initialization.}
\label{tabel:all_dataset}
\centering
\vspace{1.0 ex}
\begin{tabular}{l||ccccc|ccccc|ccccc}
\hline
\multirow{2}{*}{methods}     		   & \multicolumn{5}{c|}{Duke}                       & \multicolumn{5}{c|}{Market}                      & \multicolumn{5}{c}{CUHK03}                  \\ \cline{2-16} 
									   & rank-1 & rank-5 & rank-10 & rank-20 & mAP		 & rank-1 & rank-5 & rank-10 & rank-20 & mAP		& rank-1 & rank-5 & rank-10 & rank-20 & mAP   \\ \hline 
supervised learning                    & 63.4   & 78.9   & 83.3    & 87.4    & 42.6      & 76.2   & 89.5   & 93.3    & 95.8    & 53.1       & 24.8   & 41.1   & 52.4    & 64.2    & 23.2  \\ \hline
baseline (Duke)                        & -      & -      & -       & -       & -         & 36.1   & 52.8   & 60.9    & 69.2    & 14.2       & 4.4    & 9.9    & 13.7    & 19.4    & 4.0   \\ 
PUL (Duke)                             & -      & -      & -       & -       & -         & 44.7   & 59.1   & 65.6    & 71.7    & 20.1       & 5.6    & 11.2   & 15.8    & 22.7    & 5.2   \\ \hline
baseline (Market)                      & 21.9   & 38.3   & 44.4    & 50.5    & 10.9      & -      & -      & -       & -       & -          & 5.5    & 10.7   & 14.4    & 19.2    & 4.4   \\
PUL (Market)                           & 30.4   & 44.5   & 50.7    & 56.0    & 16.4      & -      & -      & -       & -       & -          & 7.6    & 13.8   & 18.4    & 25.1    & 7.3   \\ \hline
baseline (CUHK03)                      & 14.9   & 25.5   & 31.4    & 37.5    & 7.0       & 30.0   & 46.4   & 53.9    & 61.3    & 11.5       & -      & -      & -       & -       & -     \\
PUL (CUHK03)                           & 23.0   & 34.0   & 39.5    & 44.2    & 12.0      & 41.9   & 57.3   & 64.3    & 70.5    & 18.0       & -      & -      & -       & -       & -     \\ \hline
\end{tabular}
\end{table*}

\textbf{Market-1501} contains 32,668 images of 1,501 identities in total, which are captured from 6 cameras placed in front of a campus supermarket. The images are detected and cropped using the Deformable Part Model (DPM) \cite{DBLP:journals/pami/FelzenszwalbGMR10}. The dataset is split into three parts: 12,936 images with 751 identities for training, 19,732 images with 750 identities for gallery, and another 3,368 hand-drawn images with the same 750 gallery identities for query. All the images are of the size $128\times 64$. For simplicity, we use ``Market'' to represent this dataset.

\begin{figure}
\includegraphics[width=0.495\textwidth]{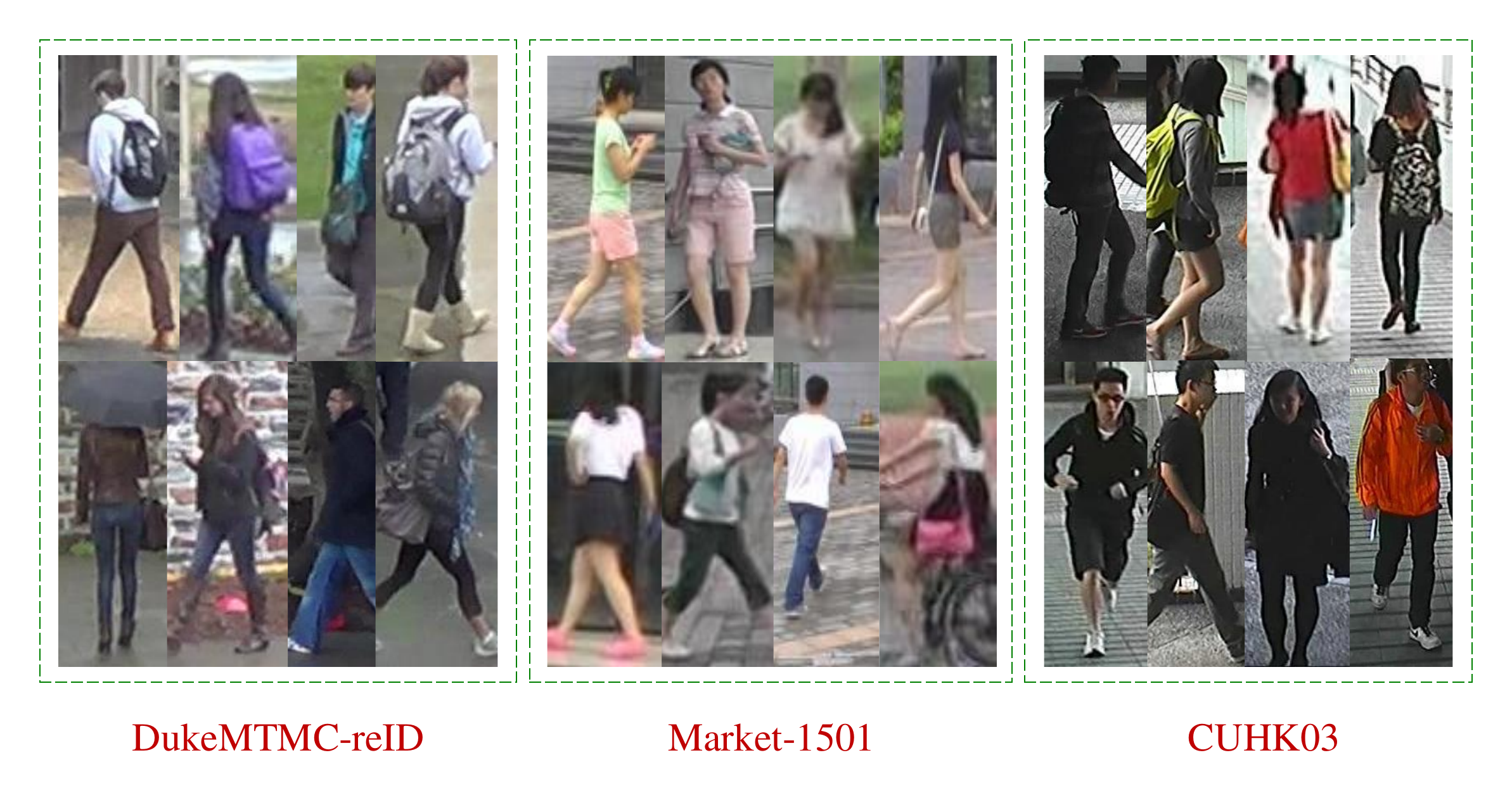}
\caption{Sample images from Duke, Market and CUHK03 datasets.}
\label{Fig:Datasets}
\end{figure}

\textbf{CUHK03} contains 14,096 images of 1,467 identities. Each identity is captured from 2 cameras in the CUHK campus, and has an average of 4.8 images in each camera. The dataset provides both manually labeled
images and DPM-detected images. We experiment on the DPM-detected images. Note that the original evaluation protocol of CUHK03 has 20 train/test splits. Nevertheless, to be in consistency with Market-1501 and Duke-reID, we use the train/test protocol proposed in \cite{DBLP:journals/corr/ZhongZCL17}: 7,365 images with 767 identities for training, 5,332 images with the remaining 700 identities for gallery, and 1,400 images with the same 700 gallery identities for query. Images in this dataset also vary in size. We list the statistics and some some samples of the these three datasets in Table \ref{tabel:dataset} and Fig \ref{Fig:Datasets}.

We report the rank-1,5,10,20 accuracy and mean average precision (mAP) for all the three datasets. All experiments use single query.

\subsection{Implementation Details}
\label{sub:implementation}

We use  ResNet-50 \cite{DBLP:conf/cvpr/HeZRS16} pre-trained on \textit{ImageNet} as our basic CNN model. To fine tune the model using the classification framework \cite{DBLP:journals/corr/ZhengYH16}, we modify the fully-connected layer to adapt to different datasets. We also insert a dropout layer before the fully-connected layer and set the dropout rate to 0.5 for all datasets. All images are  resized to 224 $\times$ 224. For data augmentation, we randomly rotate images within 20 degrees and shift them horizontally and vertically within 45 pixels. Batch size is set to 16. We use stochastic gradient descent with a momentum of 0.9 and the learning rate is set to 0.001 without any decay during the whole training process. Unless otherwise specified, ResNet-50 is fine-tuned for 20 epochs within each PUL iteration.

During feature extraction, for a given image, we use the output of the average-pooling layer as the visual representation. The $l_2$ normalized representations are used in sample selection and retrieval, while clustering uses features without normalization ($l_2$ normalization in clustering yields inferior results in our preliminary experiments). 

For the clustering step, we use \textit{k}-means++ \cite{DBLP:conf/soda/ArthurV07} to select initial cluster centers for \textit{k}-mean clustering to speed up convergence. The maximum number of iterations of the \textit{k}-means algorithm within each PUL iteration is set to 300.

\subsection{Performance of Baseline System}
\label{sub:baseline}
The basic ResNet-50 model is fine-tuned for 40 epochs on each training set. Note that our focus is not on how to improve the performance by modifying neural networks \cite{DBLP:journals/corr/ZhengZY17,sun2017svdnet} or post-processing \cite{DBLP:journals/corr/ZhongZCL17,bai2017scalable,ye2016person,zheng2015query}. Therefore, results in Table \ref{tabel:all_dataset} do not represent the state of the art.

We compare PUL with deploying the fine-tuned CNN model on an unseen testing dataset (baseline). Results are shown in Table \ref{tabel:all_dataset}.The reliability threshold $\lambda$ is set to be 0.85 for all the three datasets. Since Duke, Market and CUHK03 have 702, 751 and 767 training IDs respectively, the number of clusters $K$ is set to be 700 for Duke, and 750 for Market and CUHK03. 
When the fine-tuned model is directly deployed on other datasets, re-ID accuracy is far inferior to training/testing on the same dataset. For example, the CNN model fine-tuned and tested on Market achieves 76.2\% in rank-1 accuracy, but drops to 21.9\% when trained on Duke and tested on Market. This is because of the changes in data distribution between different datasets. In stark contrast, PUL effectively improves the re-ID accuracy over the baseline. For example, when using Duke to train the original CNN model, PUL leads to an improvement of +8.6\% in rank-1 accuracy and +5.9\% in mAP on Market-1501. 

\subsection{Evaluation of PUL}
\label{sec:evaluation_pul}

\subsubsection{Ablation study --- PUL without sample selection}
The selection of reliable samples for CNN fine-tuning is a key component in the PUL system. Here we investigate how the system works without reliable sample selection. In our experiment, for each PUL iteration, we use all the samples as training data for CNN fine-tuning. That is, after $k$-means clustering, all the images in each cluster are viewed as a class and fed into CNN.

\begin{table}[t]
\caption{Re-ID accuracy of PUL without sample selection.}
\label{table:no_selection}
\setlength{\tabcolsep}{10.0pt}
\centering
\vspace{1.0 ex}
\begin{tabular}{l|ccccc}
\hline
\# clusters ($K$) 	    & 250        & 500   	& 750 	  & 1,000    & 1,250 \\ \hline \hline
rank-1                  & 32.2       & 34.6     & 37.3    & 36.7     & 37.5  \\ 
mAP  		            & 12.8       & 14.5     & 15.7    & 14.6     & 15.4  \\ \hline
\end{tabular}
\end{table}

\begin{figure*}[t]
\label{fig6}
\centering
\subfigure[]{
        \includegraphics[width=0.230\textwidth]{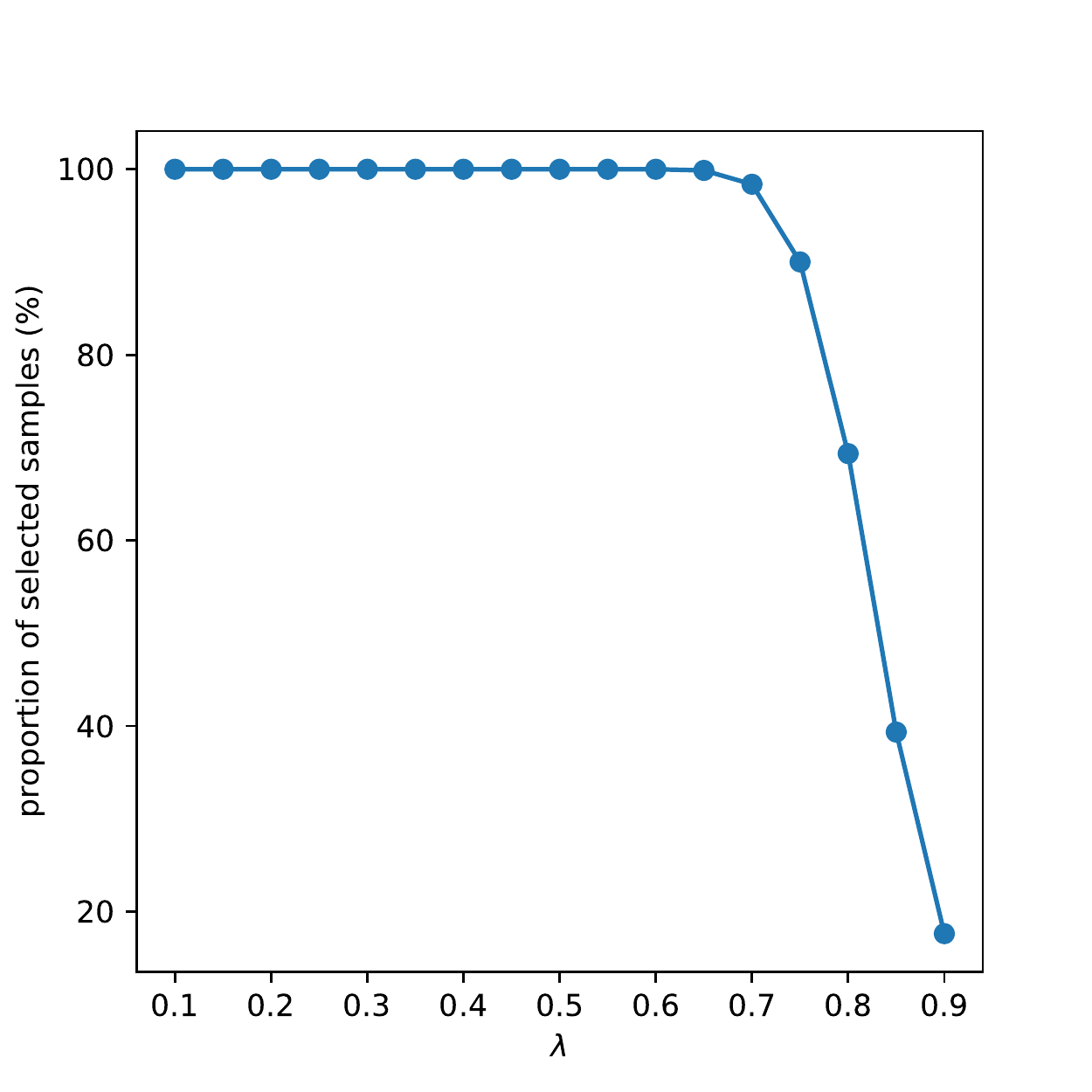}
        \label{Parameter-1}
}
\subfigure[]{
        \includegraphics[width=0.230\textwidth]{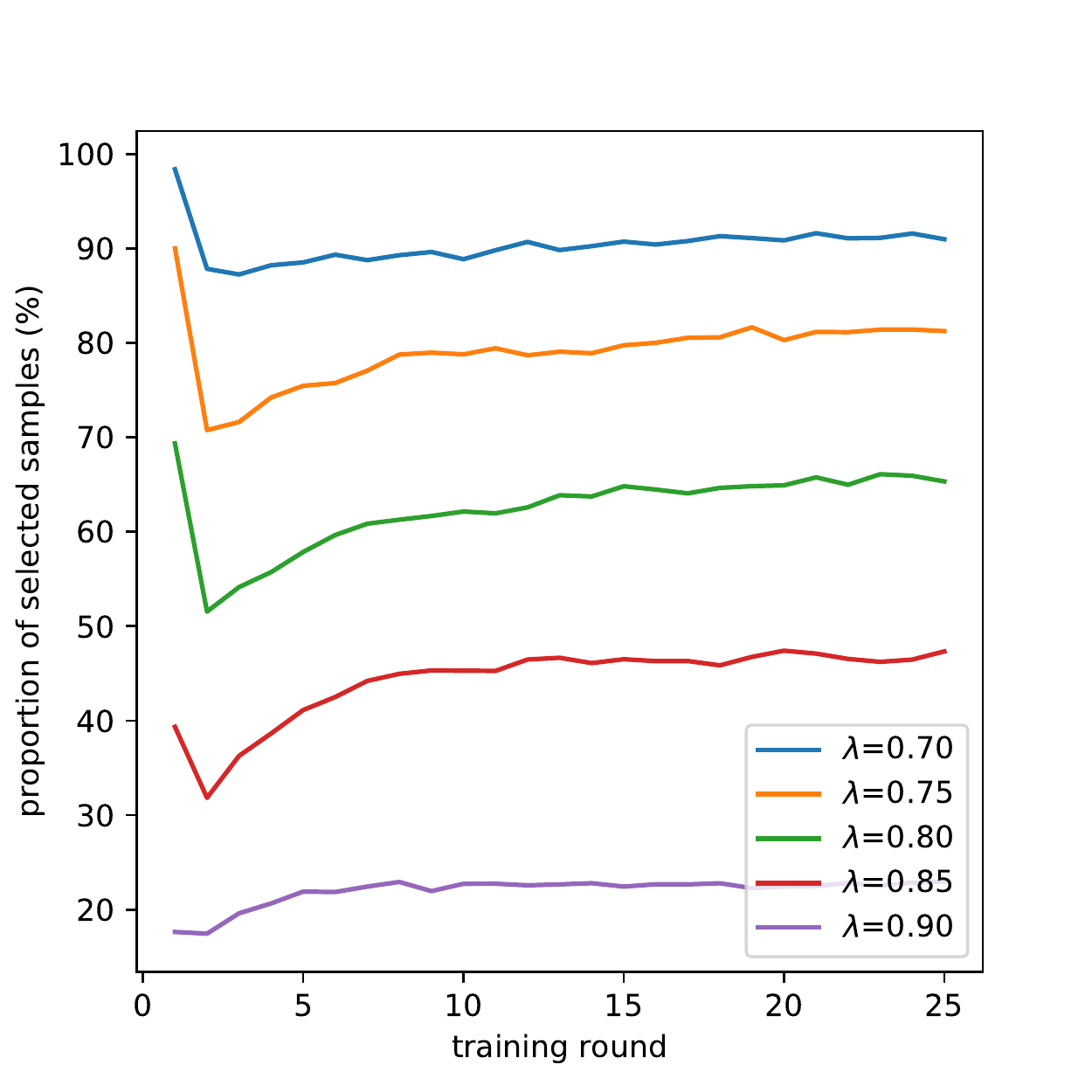}
        \label{Parameter-2}
}
\subfigure[]{
        \includegraphics[width=0.230\textwidth]{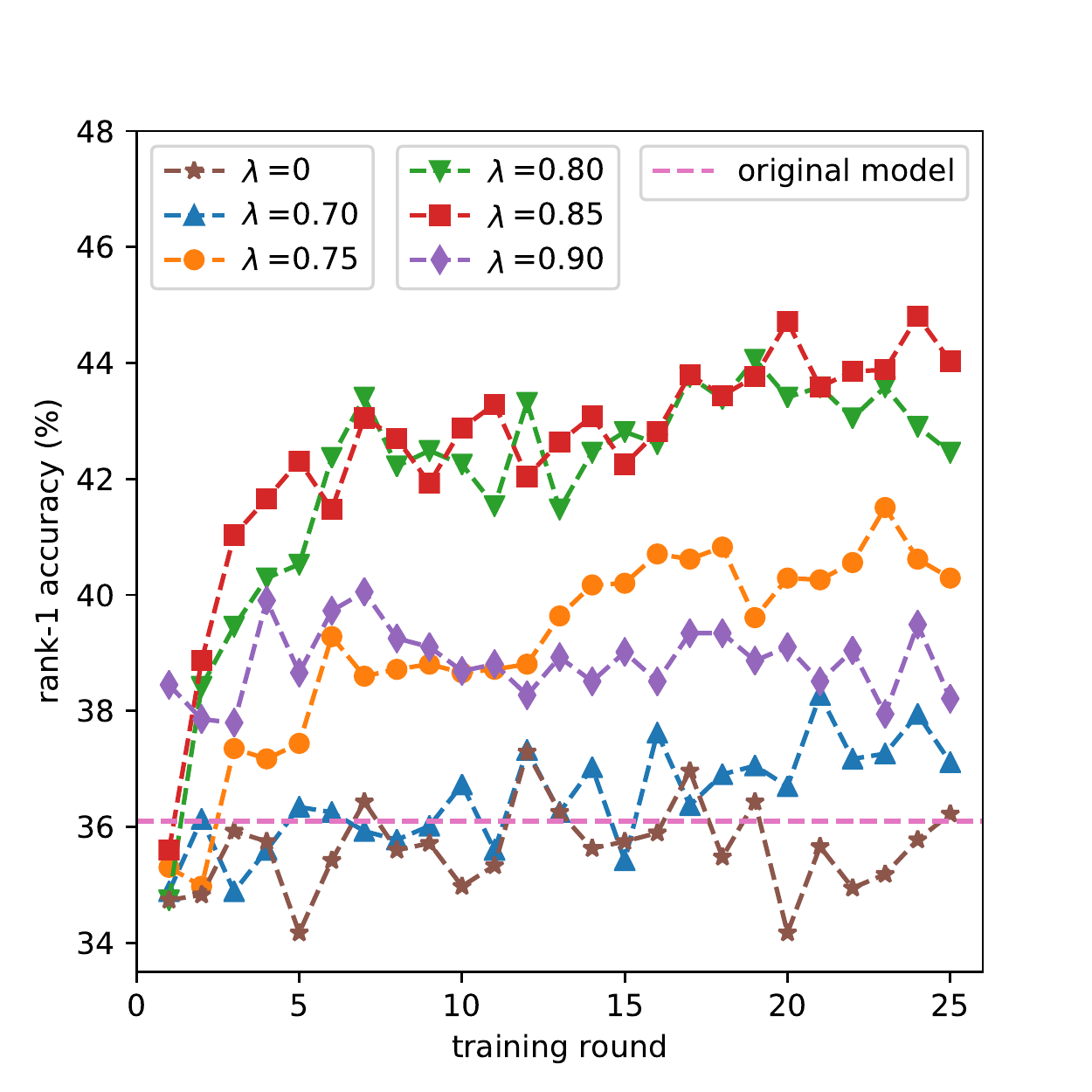}
        \label{Parameter-3}
}
\subfigure[]{
        \includegraphics[width=0.230\textwidth]{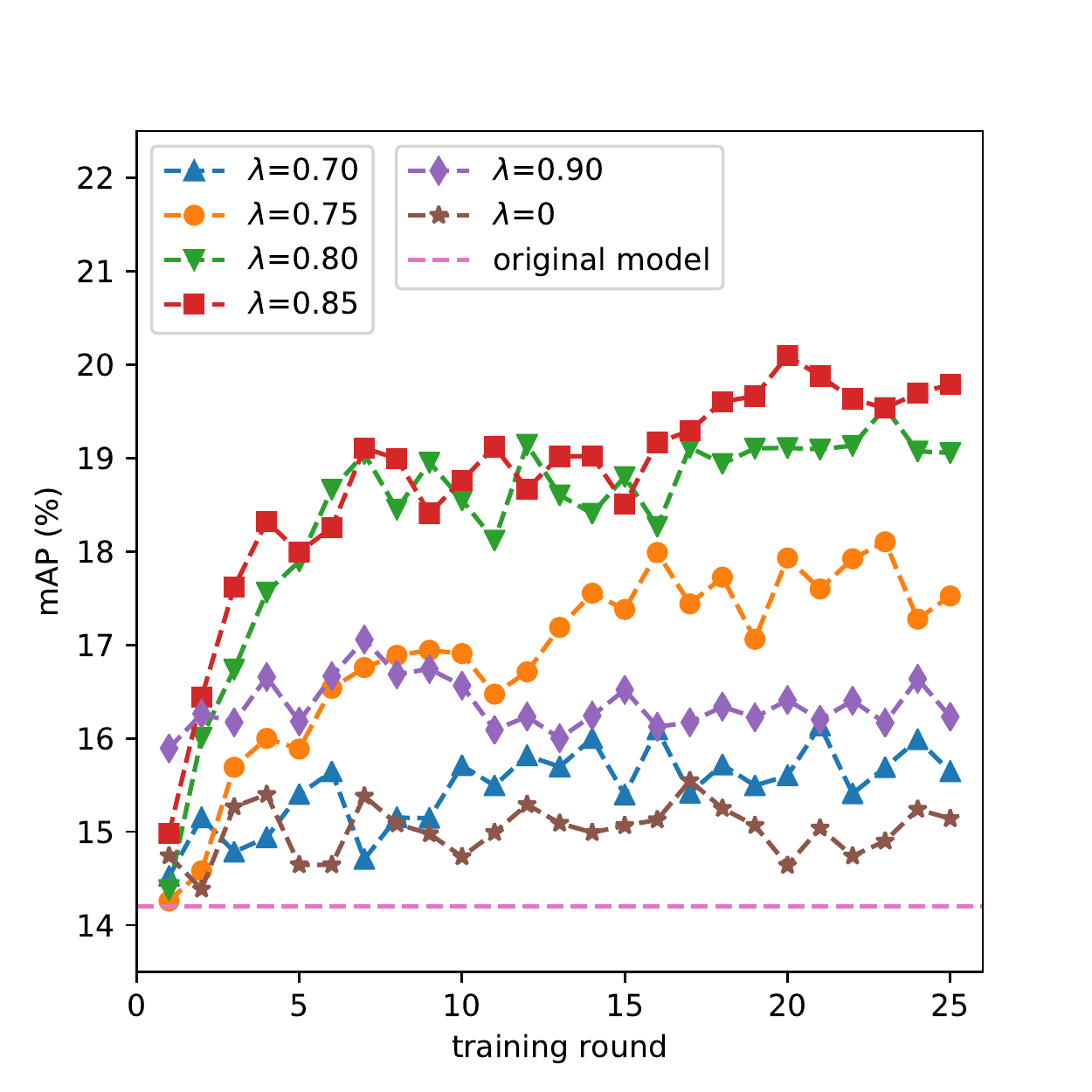}
        \label{Parameter-4}
}
\caption{The impact of $\lambda$ on re-ID accuracy ($K=750$). Since we use cosine distance, a larger $\lambda$ means a stricter sample selection process, and vice versa.  (a): proportion of selected samples in the 1st training iteration; (b): proportion of selected samples during the whole training process; (c) and (d): performance changes during whole training process. The baseline is plotted in the dashed horizontal line. }
\end{figure*}

Results are presented in Table \ref{table:no_selection}, Fig. \ref{Parameter-3} and Fig. \ref{Parameter-4}. Comparing with the baseline results in Table \ref{tabel:all_dataset}, we hardly observe any noticeable improvement of using PUL without sample selection. When $K = 1,250$, the largest gain over the baseline we can obtain is +1.4\% in rank-1 accuracy, and +1.2\% in mAP. This indicates that the samples within each cluster are very noisy, and that if we use all the noisy samples for fine-tuning without any selection, the learned CNN model gets stuck in a bad optimum. These results suggest that reliable sample selection is a necessary component in PUL.

\subsubsection{Further understanding of PUL with parameter changes}
Two important parameters are involved in this paper, \emph{i.e.,} the reliability threshold $\lambda$ and the number of clusters $K$. To this end, we mainly deploy the fine-tuned ResNet-50 model on Duke described in Section \ref{sub:baseline}. We apply PUL using this CNN model on Market-1501 without any labels. PUL is always trained for 25 iterations. 

\begin{figure}[t]
\centering
\subfigure[]{
        \includegraphics[width=0.22\textwidth]{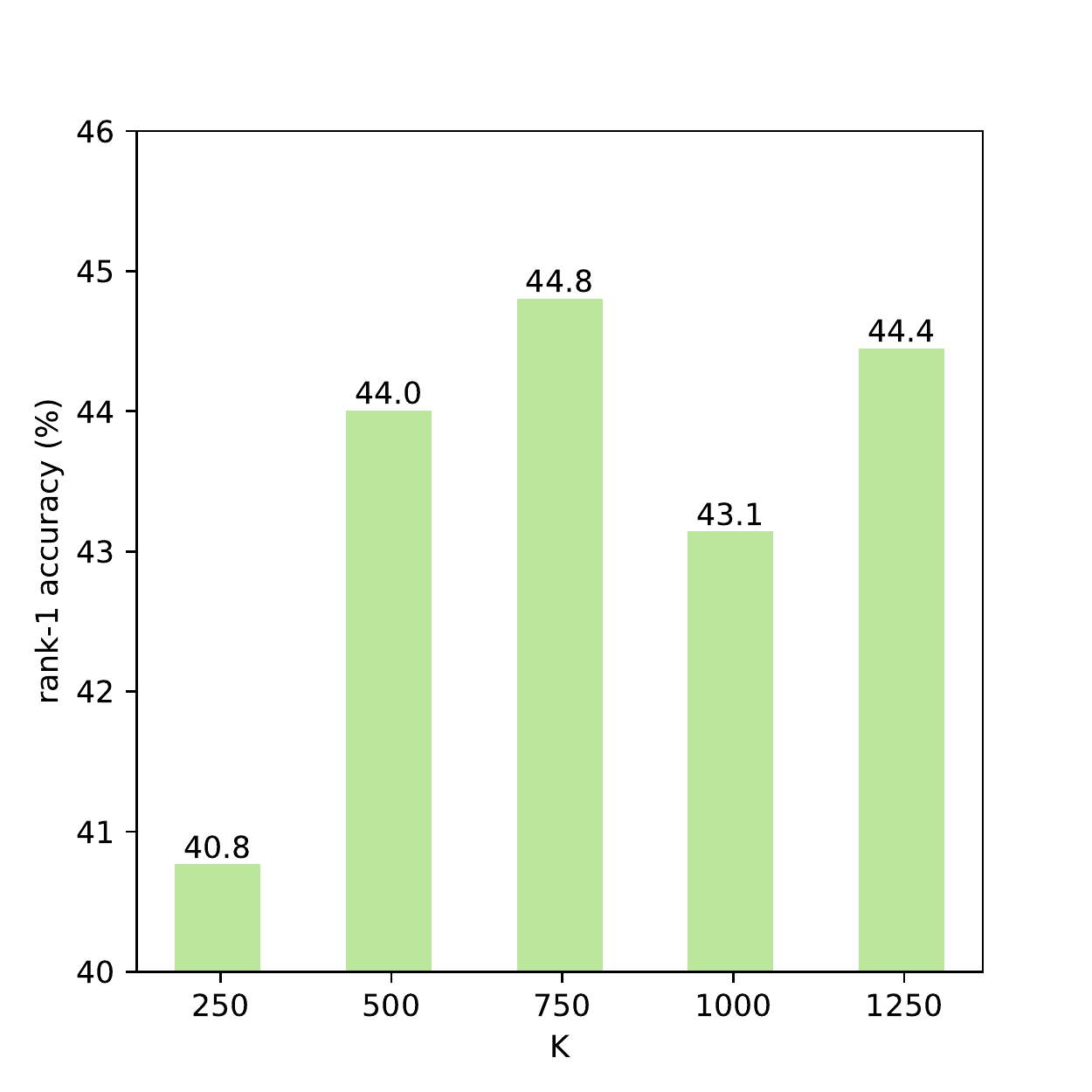}
}
\subfigure[]{
        \includegraphics[width=0.22\textwidth]{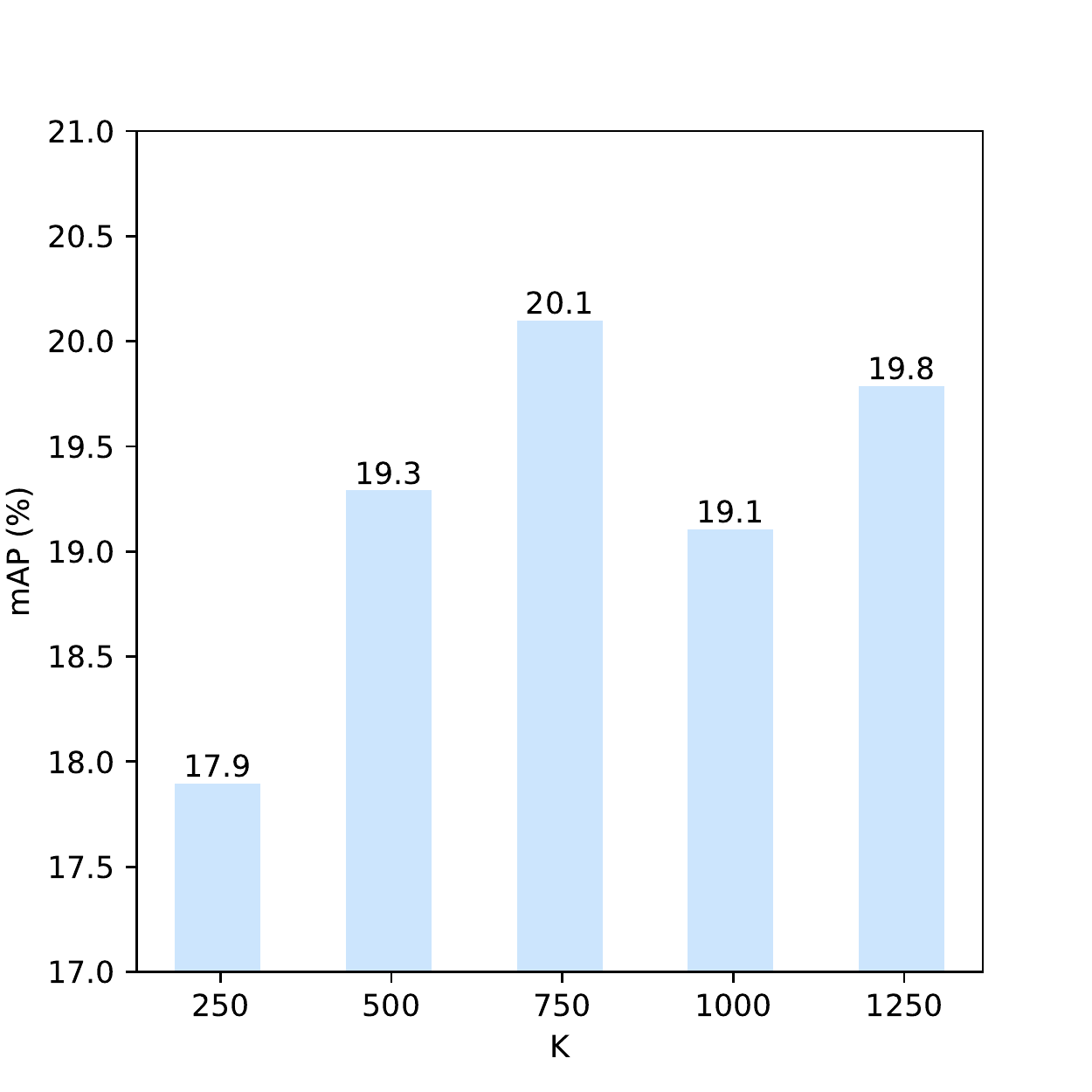}
}
\caption{Impact of the number of clusters $K$ on re-ID accuracy. ($\lambda = 0.85$). (a): rank-1 accuracy. (b): mAP.}
\label{fig:K}
\end{figure}

First, we evaluate the impact of $K$ (we fix $\lambda$ to 0.85), which is sampled from $\{250, 500, 750, 1,000, 1,250\}$, because Market has 751 training identities (not known in practice). Results are shown in Fig. \ref{fig:K}. Since the Market dataset has 751 identities, $K = 750$ achieves the best performance as expected.

Second, we evaluate the impact of $\lambda$, keeping $K=750$.
Since we use the cosine distance to measure the similarity of two feature vectors and ResNet uses ReLU (rectified linear unit) as activation function, the range of $\lambda$ should be $[0, 1]$. To further narrow the test range for $\lambda$, we present the proportions of selected samples under different $\lambda$ in the first training iteration in Fig. \ref{Parameter-1}. When $\lambda < 0.7$, nearly all of data instances are selected as being reliable, which is undesirable. Therefore, we choose $\lambda$ from $\{0.70, 0.75, 0.80, 0.85, 0.90\}$. Fig. \ref{Parameter-2} demonstrates the change of the portion of selected samples during training. As training goes on, more and more samples are selected. Let us take $\lambda = 0.85$ for instance. Before the 9th training iteration, the proportion of selected samples increases fast from 31.8\% to 45.3\%. After that, the number oscillates between 45.3\% and 47.4\%. The curve becomes saturated after 20 epochs, which indicates that PUL training can be stopped when the number of selected samples converges.

An interesting phenomenon in Fig \ref{Parameter-2} is that, when $\lambda \leq 0.85$, the proportion of reliable samples in the 1st training iteration witnesses a sudden drop after the 2nd training iteration.  This is because, in the 1st training iteration, the original model fine-tuned on Duke is too weak to discriminate the data from Market. As a result, too many samples are incorrectly selected as reliable samples. In the 2nd training iteration, the model has been fine-tuned on Market and can better separate reliable and unreliable samples on Market. 

We report the re-ID performance with different $\lambda$ in Fig \ref{Parameter-3} and Fig \ref{Parameter-4}. In general, performance improves with more iterations, indicating that the model gradually gets stronger. Specifically, $\lambda = 0.85$ yields superior accuracy. 

\subsection{Semi-supervised Re-ID}
The semi-supervised learning scheme is briefly introduced in Section \ref{sub:semi-supervised}. In this case, we initialize the CNN model with labeled images of $L$ IDs. These labeled images are also used in the subsequent CNN fine-tuning iterations. In our experiment, we set $L=25$ and $L = 50$.
We report the results in Table \ref{tabel:comparison}. Two major observations can be made.

First, comparing with the unsupervised learning method (PUL), adding 25 or 50 IDs as supervised information notably improves the re-ID accuracy. For example, when testing on Market-1501 with 25 labeled IDs, Duke and CUHK03, we observe improvement of +2.2\%, +2.7\% and +0.8\% in rank-1 accuracy, respectively. The improvement on CUHK03 is smaller compared with Market and Duke. It is probably because of the severe illumination change and less diversified training samples in CUHK03: learning an invariant feature is extremely difficult.

Second, we observe that using more IDs as training samples is beneficial to the system. For example, when testing on Market-1501, using 50 IDs brings about +3.2\% more improvement than using 25 IDs. On the other two testing sets, the additional improvement brought by the 50 IDs is +3.8\% and +0.2\% in rank-1 accuracy, respectively. These results suggest that using more labeled data will increase the model performance at the cost of higher labeling cost.

\subsection{Comparison with Non-deep Methods}
\label{sub:comparison}

\begin{table*}[t]
\caption{Person re-ID accuracy (rank-1,5,10,20 precision and mAP) of five models: BoW, LOMO, UMDL, the baseline (fine-tuned ResNet-50 on two of the three datasets and tested on the last one), and PUL (both the unsupervised and the semi-supervised versions are shown). }
\label{tabel:comparison}
\setlength{\tabcolsep}{3.05pt}
\centering
\vspace{1.0 ex}
\begin{tabular}{lll||ccccc|ccccc|ccccc}
\hline
\multicolumn{3}{c||}{\multirow{2}{*}{methods}}     		     & \multicolumn{5}{c|}{Duke, CUHK03 $\rightarrow$ Market}          & \multicolumn{5}{c|}{Market, CUHK03 $\rightarrow$ Duke}     & \multicolumn{5}{c}{Duke, Market $\rightarrow$ CUHK03}  \\ \cline{4-18} 

\multicolumn{3}{c||}{}									     & rank-1 & rank-5 & rank-10 & rank-20 & mAP		               & rank-1 & rank-5 & rank-10 & rank-20 & mAP		            & rank-1 & rank-5 & rank-10 & rank-20 & mAP   \\ \hline 

\multirow{5}{*}{\rotatebox[origin=c]{90}{unsupervised}} & \multirow{2}{*}{\rotatebox[origin=c]{90}{hand}}  & BoW\cite{DBLP:conf/iccv/ZhengSTWWT15}   & 35.8   & 52.4   & 60.3    & 67.6    & 14.8                      & 17.1   & 28.8   & 34.9    & 41.2    & 8.3                  & 2.1    & 4.6    & 7.0     & 11.2    & 1.9   \\ 

                                                        &                                                  & LOMO\cite{DBLP:conf/cvpr/LiaoHZL15}     & 27.2   & 41.6   & 49.1    & 57.2    & 8.0                       & 12.3   & 21.3   & 26.6    & 33.3    & 4.8                  & 0.6    & 1.9    & 3.6     & 5.4     & 0.7   \\ \cline{2-18} 
                                                        
                                                        & \multicolumn{2}{l||}{UMDL\cite{peng2016unsupervised}}                                       & 34.5   & 52.6   & 59.6    & 68.0    & 12.4                      & 18.5   & 31.4   & 37.6    & 44.7    & 7.3                  & 1.4    & 5.0    & 7.1     & 9.7     & 1.3   \\ 
                                                       
                                                        & \multicolumn{2}{l||}{Deep feature (baseline)}                                               & 40.0   & 56.5   & 63.7    & 70.2    & 17.0                      & 22.9   & 37.3   & 44.2    & 50.8    & 11.3                 & 4.3    & 10.1   & 15.1    & 19.4    & 3.9   \\ 
                                                        
                                                        & \multicolumn{2}{l||}{Deep feature (PUL)}                                                    & 45.5   & 60.7   & 66.7    & 72.6    & 20.5                      & 30.0   & 43.4   & 48.5    & 54.0    & 16.4                 & 8.1    & 15.8   & 22.0    & 28.5    & 7.9   \\ \hline
                                                        
\multirow{4}{*}{\rotatebox[origin=c]{90}{semi-sup}}     & \multirow{2}{*}{\rotatebox[origin=c]{90}{25ID}}  & UMDL\cite{peng2016unsupervised}          & 35.2   & 53.2   & 60.2    & 68.4    & 13.3                      & 18.9   & 31.7   & 37.8    & 45.2    & 7.6                  & 1.4    & 5.2    & 7.2     & 10.0     & 1.3   \\ 

                                                        &                                                  & Deep feature (PUL)                       & 47.7   & 63.6   & 69.3    & 75.5    & 22.2                      & 32.7   & 47.2   & 53.0    & 59.1    & 18.8                 & 8.9    & 18.4   & 23.8    & 28.5    & 8.8   \\ \cline{2-18}
                                                        
                                                        & \multirow{2}{*}{\rotatebox[origin=c]{90}{50ID}}  & UMDL\cite{peng2016unsupervised}          & 35.6   & 53.5   & 60.4    & 68.6    & 13.4                      & 19.5   & 32.4   & 38.6    & 45.7    & 8.3                  & 1.6    & 5.4    & 7.5     & 10.2     & 1.4   \\ 
                                                        &                                                  & Deep feature (PUL)                       & 50.9   & 66.5   & 72.6    & 78.3    & 24.8                      & 36.5   & 52.6   & 57.9    & 64.5    & 21.5                 & 9.1    & 18.6   & 25.1    & 32.5    & 9.2   \\ \hline
\end{tabular}
\end{table*}

We then compare the PUL method with competing unsupervised methods. As we mentioned in Section \ref{sub:related-unsupervised}, limited attention has been paid to unsupervised learning for re-ID and fewer works have adopted deep features. Since these unsupervised methods mainly focus on re-ID on relatively small datasets, benchmarking results are lacking on larger datasets such as Market. In this section, we evaluate the performances of two hand-crafted features, i.e., Bag-of-Words (BoW) \cite{DBLP:conf/iccv/ZhengSTWWT15}, local maximal occurrence (LOMO) \cite{DBLP:conf/cvpr/LiaoHZL15} and one transfer learning method, i.e., the Unsupervised Multi-task Dictionary Learning (UMDL) model \cite{peng2016unsupervised} on the large-scale datasets. To our knowledge, except hand-crafted features, UMDL is the only work on unsupervised re-ID whose code is released. 

Note that, for BoW and LOMO, there is no training process. We directly apply these two hand-crafted features on the test datasets. After feature extraction, we normalize the feature vectors and use Euclidean distance to measure the similarity.

UMDL needs multiple datasets to learning a task-independent dictionary. To meet this requirement, we use two of the three datasets as labeled datasets and treat the remaining one as an unlabeled dataset. Since the released code of UMDL does not include feature extraction, we use the BoW feature proposed in \cite{DBLP:conf/iccv/ZhengSTWWT15} for this method. Since the computational complexity of UMDL is prohibitively high, we choose 2,500 images from each dataset in training. For PUL, we still fix the reliability threshold $\lambda$ to 0.85 for all the three datasets. The number of clusters $K$ is set to 750 for Market and CUHK03, and 700 for Duke. Results are presented in Table \ref{tabel:comparison}, from which two conclusions can be drawn.

First, when comparing Table \ref{tabel:all_dataset} and Table \ref{tabel:comparison}, we find that initialization using more labeled datasets does not  noticeably improve re-ID accuracy. Sometimes, multi-dataset initialization yields even worse results than single dataset initialization. For example, when tested on duke,  PUL achieves 30.0\% in rank-1 accuracy when initialized by both Market+CUHK03 but 30.4\% in rank-1 accuracy when initialized only by Market. The baseline approach also exhibits similar phenomenon when tested on CUHK03. This indicates that using more labeled datasets may be not a best choice for unsupervised re-ID. In comparison, PUL with unlabeled data can always improve the re-ID accuracy, regardless of the datasets used for CNN initialization.   

Second, both the CNN baseline and PUL outperform UMDL by a large margin. For example, when tested on three datasets, the CNN baseline outperforms UMDL by +5.5\%, +4.4\%, and +2.9\% in rank-1 accuracy, respectively. When PUL is performed, we observe larger performance gaps, arriving at +8.7\%, +11.5\% and +6.7\% in rank-1 accuracy on the three datasets, respectively. Under the semi-supervised setting (Section \ref{sub:semi-supervised}), the superiority of PUL is even more significant. For example, when using 50 labeled IDs, PUL outperforms UMDL by +15.3\%, +17.0\%, and +7.5\% in rank-1 accuracy on the three datasets, respectively. In fact, as more data is labeled, UMDL does not exhibit noticeable improvement, while PUL shows clear performance gain. Moreover, considering the training inefficiency of UMDL, we can thus validate the effectiveness of PUL.

\section{Conclusion}
In this paper, a simple progressive unsupervised learning (PUL) method is proposed for person re-ID, based on the iterations between $k$-means clustering and CNN fine-tuning. We show that reliable sample selection is a key component to its success. Progressively, the proposed method produces CNN models with high discriminative ability. 

The proposed method is easy to implement and a number of further extensions can be made. For example, as mentioned in \cite{DBLP:journals/corr/ZhengYH16}, in video re-ID, frames within a tracklet can be viewed as belonging to the same ID, which can be used to initialize clustering and fine-tuning. Another promising idea is to integrate diversity into PUL, so that training samples from multiple cameras can be selected for fine-tuning.


%





\ifCLASSOPTIONcaptionsoff
  \newpage
\fi



\bibliographystyle{IEEEtran}
\bibliography{IEEEabrv,egbib}

\begin{thebibliography}{10}
\providecommand{\url}[1]{#1}
\csname url@samestyle\endcsname
\providecommand{\newblock}{\relax}
\providecommand{\bibinfo}[2]{#2}
\providecommand{\BIBentrySTDinterwordspacing}{\spaceskip=0pt\relax}
\providecommand{\BIBentryALTinterwordstretchfactor}{4}
\providecommand{\BIBentryALTinterwordspacing}{\spaceskip=\fontdimen2\font plus
\BIBentryALTinterwordstretchfactor\fontdimen3\font minus
  \fontdimen4\font\relax}
\providecommand{\BIBforeignlanguage}[2]{{%
\expandafter\ifx\csname l@#1\endcsname\relax
\typeout{** WARNING: IEEEtran.bst: No hyphenation pattern has been}%
\typeout{** loaded for the language `#1'. Using the pattern for}%
\typeout{** the default language instead.}%
\else
\language=\csname l@#1\endcsname
\fi
#2}}
\providecommand{\BIBdecl}{\relax}
\BIBdecl

\bibitem{DBLP:journals/corr/ZhengYH16}
L.~Zheng, Y.~Yang, and A.~G. Hauptmann, ``Person re-identification: Past,
  present and future,'' \emph{CoRR}, 2016.

\bibitem{DBLP:conf/iccv/ZhengSTWWT15}
L.~Zheng, L.~Shen, L.~Tian, S.~Wang, J.~Wang, and Q.~Tian, ``Scalable person
  re-identification: {A} benchmark,'' in \emph{ICCV}, 2015.

\bibitem{DBLP:journals/corr/ZhengZY17}
Z.~Zheng, L.~Zheng, and Y.~Yang, ``Unlabeled samples generated by {GAN} improve
  the person re-identification baseline in vitro,'' \emph{CoRR}, 2017.

\bibitem{varior2016gated}
R.~R. Varior, M.~Haloi, and G.~Wang, ``Gated siamese convolutional neural
  network architecture for human re-identification,'' in \emph{ECCV}, 2016.

\bibitem{liu2016end}
H.~Liu, J.~Feng, M.~Qi, J.~Jiang, and S.~Yan, ``End-to-end comparative
  attention networks for person re-identification,'' \emph{CoRR}, 2016.

\bibitem{chen2017beyond}
W.~Chen, X.~Chen, J.~Zhang, and K.~Huang, ``Beyond triplet loss: a deep
  quadruplet network for person re-identification,'' in \emph{CVPR}, 2017.

\bibitem{cheng2016person}
D.~Cheng, Y.~Gong, S.~Zhou, J.~Wang, and N.~Zheng, ``Person re-identification
  by multi-channel parts-based cnn with improved triplet loss function,'' in
  \emph{CVPR}, 2016.

\bibitem{DBLP:journals/corr/HermansBL17}
A.~Hermans, L.~Beyer, and B.~Leibe, ``In defense of the triplet loss for person
  re-identification,'' \emph{CoRR}, 2017.

\bibitem{xiao2016learning}
T.~Xiao, H.~Li, W.~Ouyang, and X.~Wang, ``Learning deep feature representations
  with domain guided dropout for person re-identification,'' in \emph{CVPR},
  2016.

\bibitem{DBLP:journals/corr/LinZZWY17}
Y.~Lin, L.~Zheng, Z.~Zheng, Y.~Wu, and Y.~Yang, ``Improving person
  re-identification by attribute and identity learning,'' \emph{CoRR}, 2017.

\bibitem{DBLP:conf/eccv/KodirovXFG16}
E.~Kodirov, T.~Xiang, Z.~Fu, and S.~Gong, ``Person re-identification by
  unsupervised $l_1$ graph learning,'' in \emph{ECCV}, 2016.

\bibitem{peng2016unsupervised}
P.~Peng, T.~Xiang, Y.~Wang, M.~Pontil, S.~Gong, T.~Huang, and Y.~Tian,
  ``Unsupervised cross-dataset transfer learning for person
  re-identification,'' in \emph{CVPR}, 2016.

\bibitem{ma2017person}
X.~Ma, X.~Zhu, S.~Gong, X.~Xie, J.~Hu, K.-M. Lam, and Y.~Zhong, ``Person
  re-identification by unsupervised video matching,'' \emph{Pattern
  Recognition}, 2017.

\bibitem{yang2017unsupervised}
Y.~Yang, S.~Li, L.~Wen, and S.~Lyu, ``Unsupervised learning of multi-level
  descriptors for person re-identification,'' in \emph{AAAI}, 2017.

\bibitem{DBLP:journals/corr/YangWHTR17}
X.~Yang, M.~Wang, R.~Hong, Q.~Tian, and Y.~Rui, ``Enhancing person
  re-identification in a self-trained subspace,'' \emph{ACM Transactions on
  Multimedia Computing, Communications, and Applications}, 2017.

\bibitem{gray2007evaluating}
D.~Gray, S.~Brennan, and H.~Tao, ``Evaluating appearance models for
  recognition, reacquisition, and tracking,'' in \emph{Proc. IEEE International
  Workshop on Performance Evaluation for Tracking and Surveillance (PETS)},
  2007.

\bibitem{wang2014person}
T.~Wang, S.~Gong, X.~Zhu, and S.~Wang, ``Person re-identification by video
  ranking,'' in \emph{ECCV}, 2014.

\bibitem{sun2017svdnet}
Y.~Sun, L.~Zheng, W.~Deng, and S.~Wang, ``Svdnet for pedestrian retrieval,''
  \emph{arXiv preprint arXiv:1703.05693}, 2017.

\bibitem{geng2016deep}
M.~Geng, Y.~Wang, T.~Xiang, and Y.~Tian, ``Deep transfer learning for person
  re-identification,'' \emph{arXiv preprint arXiv:1611.05244}, 2016.

\bibitem{wang2016human}
H.~Wang, S.~Gong, X.~Zhu, and T.~Xiang, ``Human-in-the-loop person
  re-identification,'' in \emph{ECCV}, 2016.

\bibitem{liu2013pop}
C.~Liu, C.~Change~Loy, S.~Gong, and G.~Wang, ``Pop: Person re-identification
  post-rank optimisation,'' in \emph{ICCV}, 2013.

\bibitem{DBLP:conf/nips/KumarPK10}
M.~P. Kumar, B.~Packer, and D.~Koller, ``Self-paced learning for latent
  variable models,'' in \emph{NIPS}, 2010.

\bibitem{DBLP:conf/nips/KrizhevskySH12}
A.~Krizhevsky, I.~Sutskever, and G.~E. Hinton, ``Imagenet classification with
  deep convolutional neural networks,'' in \emph{NIPS}, 2012.

\bibitem{DBLP:conf/icpr/YiLLL14}
D.~Yi, Z.~Lei, S.~Liao, and S.~Z. Li, ``Deep metric learning for person
  re-identification,'' in \emph{ICPR}, 2014.

\bibitem{DBLP:conf/cvpr/LiZXW14}
W.~Li, R.~Zhao, T.~Xiao, and X.~Wang, ``Deepreid: Deep filter pairing neural
  network for person re-identification,'' in \emph{CVPR}, 2014.

\bibitem{DBLP:conf/eccv/RadenovicTC16}
F.~Radenovic, G.~Tolias, and O.~Chum, ``{CNN} image retrieval learns from bow:
  Unsupervised fine-tuning with hard examples,'' in \emph{ECCV}, 2016.

\bibitem{DBLP:conf/cvpr/SchroffKP15}
F.~Schroff, D.~Kalenichenko, and J.~Philbin, ``Facenet: {A} unified embedding
  for face recognition and clustering,'' in \emph{CVPR}, 2015.

\bibitem{DBLP:conf/eccv/GrayT08}
D.~Gray and H.~Tao, ``Viewpoint invariant pedestrian recognition with an
  ensemble of localized features,'' in \emph{ECCV}, 2008.

\bibitem{DBLP:conf/eccv/ZhengBSWSWT16}
L.~Zheng, Z.~Bie, Y.~Sun, J.~Wang, C.~Su, S.~Wang, and Q.~Tian, ``{MARS:} {A}
  video benchmark for large-scale person re-identification,'' in \emph{ECCV},
  2016.

\bibitem{zheng2017pose}
L.~Zheng, Y.~Huang, H.~Lu, and Y.~Yang, ``Pose invariant embedding for deep
  person re-identification,'' \emph{arXiv preprint arXiv:1701.07732}, 2017.

\bibitem{xiao2017joint}
T.~Xiao, S.~Li, B.~Wang, L.~Lin, and X.~Wang, ``Joint detection and
  identification feature learning for person search,'' \emph{arXiv preprint
  arXiv:1604.01850}, 2017.

\bibitem{wang2016towards}
H.~Wang, X.~Zhu, T.~Xiang, and S.~Gong, ``Towards unsupervised open-set person
  re-identification,'' in \emph{ICIP}, 2016.

\bibitem{DBLP:conf/cvpr/FarenzenaBPMC10}
M.~Farenzena, L.~Bazzani, A.~Perina, V.~Murino, and M.~Cristani, ``Person
  re-identification by symmetry-driven accumulation of local features,'' in
  \emph{CVPR}, 2010.

\bibitem{DBLP:conf/cvpr/ZhaoOW13}
R.~Zhao, W.~Ouyang, and X.~Wang, ``Unsupervised salience learning for person
  re-identification,'' in \emph{CVPR}, 2013.

\bibitem{DBLP:conf/iccv/ZhaoOW13}
------, ``Person re-identification by salience matching,'' in \emph{ICCV},
  2013.

\bibitem{DBLP:conf/cvpr/ZhaoOW14}
------, ``Learning mid-level filters for person re-identification,'' in
  \emph{CVPR}, 2014.

\bibitem{DBLP:conf/cvpr/LiaoHZL15}
S.~Liao, Y.~Hu, X.~Zhu, and S.~Z. Li, ``Person re-identification by local
  maximal occurrence representation and metric learning,'' in \emph{CVPR},
  2015.

\bibitem{DBLP:conf/cvpr/ZhangXG16}
L.~Zhang, T.~Xiang, and S.~Gong, ``Learning a discriminative null space for
  person re-identification,'' in \emph{CVPR}, 2016.

\bibitem{DBLP:conf/cvpr/ZhangLLIR16}
Y.~Zhang, B.~Li, H.~Lu, A.~Irie, and X.~Ruan, ``Sample-specific {SVM} learning
  for person re-identification,'' in \emph{CVPR}, 2016.

\bibitem{DBLP:conf/cvpr/ChenYCZ16}
D.~Chen, Z.~Yuan, B.~Chen, and N.~Zheng, ``Similarity learning with spatial
  constraints for person re-identification,'' in \emph{CVPR}, 2016.

\bibitem{DBLP:conf/icml/BengioLCW09}
Y.~Bengio, J.~Louradour, R.~Collobert, and J.~Weston, ``Curriculum learning,''
  in \emph{ICML}, 2009.

\bibitem{DBLP:conf/nips/JiangMYLSH14}
L.~Jiang, D.~Meng, S.~Yu, Z.~Lan, S.~Shan, and A.~G. Hauptmann, ``Self-paced
  learning with diversity,'' in \emph{NIPS}, 2014.

\bibitem{ma2017spacotrain}
M.~Fan, M.~Deyu, X.~Qi, Z.~Li, and X.~Dong, ``Self-paced cotraining,'' in
  \emph{ICML}, 2017.

\bibitem{DBLP:journals/pami/ZhangMH17}
D.~Zhang, D.~Meng, and J.~Han, ``Co-saliency detection via a self-paced
  multiple-instance learning framework,'' \emph{{IEEE} Trans. Pattern Anal.
  Mach. Intell.}, vol.~39, no.~5, pp. 865--878, 2017.

\bibitem{DBLP:conf/cvpr/HeZRS16}
K.~He, X.~Zhang, S.~Ren, and J.~Sun, ``Deep residual learning for image
  recognition,'' in \emph{CVPR}, 2016.

\bibitem{DBLP:conf/eccv/RistaniSZCT16}
E.~Ristani, F.~Solera, R.~S. Zou, R.~Cucchiara, and C.~Tomasi, ``Performance
  measures and a data set for multi-target, multi-camera tracking,'' in
  \emph{ECCV}, 2016.

\bibitem{DBLP:journals/corr/ZhongZCL17}
Z.~Zhong, L.~Zheng, D.~Cao, and S.~Li, ``Re-ranking person re-identification
  with k-reciprocal encoding,'' in \emph{CVPR}, 2017.

\bibitem{DBLP:journals/pami/FelzenszwalbGMR10}
P.~F. Felzenszwalb, R.~B. Girshick, D.~A. McAllester, and D.~Ramanan, ``Object
  detection with discriminatively trained part-based models,'' \emph{TPAMI},
  vol.~32, no.~9, 2010.

\bibitem{DBLP:conf/soda/ArthurV07}
D.~Arthur and S.~Vassilvitskii, ``k-means++: the advantages of careful
  seeding,'' in \emph{SODA}, 2007.

\bibitem{bai2017scalable}
S.~Bai, X.~Bai, and Q.~Tian, ``Scalable person re-identification on supervised
  smoothed manifold,'' in \emph{CVPR}, 2017.

\bibitem{ye2016person}
M.~Ye, C.~Liang, Y.~Yu, Z.~Wang, Q.~Leng, C.~Xiao, J.~Chen, and R.~Hu, ``Person
  reidentification via ranking aggregation of similarity pulling and
  dissimilarity pushing,'' \emph{IEEE Transactions on Multimedia}, vol.~18,
  no.~12, pp. 2553--2566, 2016.

\bibitem{zheng2015query}
L.~Zheng, S.~Wang, L.~Tian, F.~He, Z.~Liu, and Q.~Tian, ``Query-adaptive late
  fusion for image search and person re-identification,'' in \emph{CVPR}, 2015.

\end{thebibliography}

\end{document}